%% file: main.tex
\documentclass[sigconf]{acmart}

\usepackage{fontawesome}

\usepackage{multirow} 
\usepackage{hyperref}
\usepackage{bbm}

\usepackage{algorithm}
\usepackage{algorithmic}
\usepackage{amsmath}
\usepackage{xcolor}
\usepackage{booktabs}
\usepackage{graphicx}

\usepackage{amssymb}
\usepackage{colortbl}
\usepackage{array}
\usepackage{makecell}
\usepackage{float}
\usepackage{pifont}
\usepackage{subcaption}

\definecolor{purple}{HTML}{6D2382} 

\AtBeginDocument{%
  }

\setcopyright{acmlicensed}
\copyrightyear{2025}
\acmYear{2025}
\acmConference[MM '25] {Proceedings of the 33rd ACM International Conference on Multimedia}{October 27--31, 2025}{Dublin, Ireland.}
\acmBooktitle{Proceedings of the 33rd ACM International Conference on Multimedia (MM '25), October 27--31, 2025, Dublin, Ireland}
\acmISBN{979-8-4007-2035-2/2025/10}
\acmDOI{10.1145/3746027.3754861}

\begin{document}

\title{Dome-DETR: DETR with Density-Oriented Feature-Query Manipulation for Efficient Tiny Object Detection}

\author{Zhangchi Hu}
\email{huzhangchi@mail.ustc.edu.cn}
\affiliation{%
  \institution{University of Science and Technology of China}
  \city{Hefei}
  \state{Anhui}
  \country{China}
}

\author{Peixi Wu}
\email{wupeixi@mail.ustc.edu.cn}
\affiliation{%
  \institution{University of Science and Technology of China}
  \city{Hefei}
  \state{Anhui}
  \country{China}}

\author{Jie Chen}
\email{chenjie02@mail.ustc.edu.cn}
\affiliation{%
  \institution{University of Science and Technology of China}
  \city{Hefei}
  \state{Anhui}
  \country{China}}

\author{Huyue Zhu}
\email{huyuezhu@mail.ustc.edu.cn}
\affiliation{%
  \institution{University of Science and Technology of China}
  \city{Hefei}
  \state{Anhui}
  \country{China}}

\author{Yijun Wang}
\email{wangyijun@mail.ustc.edu.cn}
\affiliation{%
  \institution{University of Science and Technology of China}
  \city{Hefei}
  \state{Anhui}
  \country{China}}

\author{Yansong Peng}
\email{pengyansong@mail.ustc.edu.cn}
\affiliation{%
  \institution{University of Science and Technology of China}
  \city{Hefei}
  \state{Anhui}
  \country{China}}

\author{Hebei Li}
\email{lihebei@mail.ustc.edu.cn}
\authornote{Corresponding author.}
\affiliation{%
  \institution{University of Science and Technology of China}
  \city{Hefei}
  \state{Anhui}
  \country{China}}

\author{Xiaoyan Sun}
\email{sunxiaoyan@ustc.edu.cn}
\affiliation{%
  \institution{University of Science and Technology of China}
  \institution{Institute of Artificial Intelligence, Hefei Comprehensive National Science Center}
  \city{Hefei}
  \state{Anhui}
  \country{China}}

\renewcommand{\shortauthors}{Zhangchi Hu et al.}

\begin{abstract}
Tiny object detection plays a vital role in drone surveillance, remote sensing, and autonomous systems, enabling the identification of small targets across vast landscapes. However, existing methods suffer from inefficient feature leverage and high computational costs due to redundant feature processing and rigid query allocation. To address these challenges, we propose \textbf{Dome-DETR}, a novel framework with \textbf{D}ensity-\textbf{O}riented Feature-Query \textbf{M}anipulation for \textbf{E}fficient Tiny Object Detection. To reduce feature redundancies, we introduce a lightweight Density-Focal Extractor (DeFE) to produce clustered compact foreground masks. Leveraging these masks, we incorporate Masked Window Attention Sparsification (MWAS) to focus computational resources on the most informative regions via sparse attention. Besides, we propose Progressive Adaptive Query Initialization (PAQI), which adaptively modulates query density across spatial areas for better query allocation. Extensive experiments demonstrate that Dome-DETR achieves state-of-the-art performance \textbf{(+3.3 AP on AI-TOD-V2 and +2.5 AP on VisDrone)} while maintaining low computational complexity and a compact model size. Code is available at \href{https://github.com/RicePasteM/Dome-DETR}{https://github.com/RicePasteM/Dome-DETR}.
\end{abstract}




\maketitle

\begin{figure}[t]
  \centering
  \includegraphics[width=\linewidth]{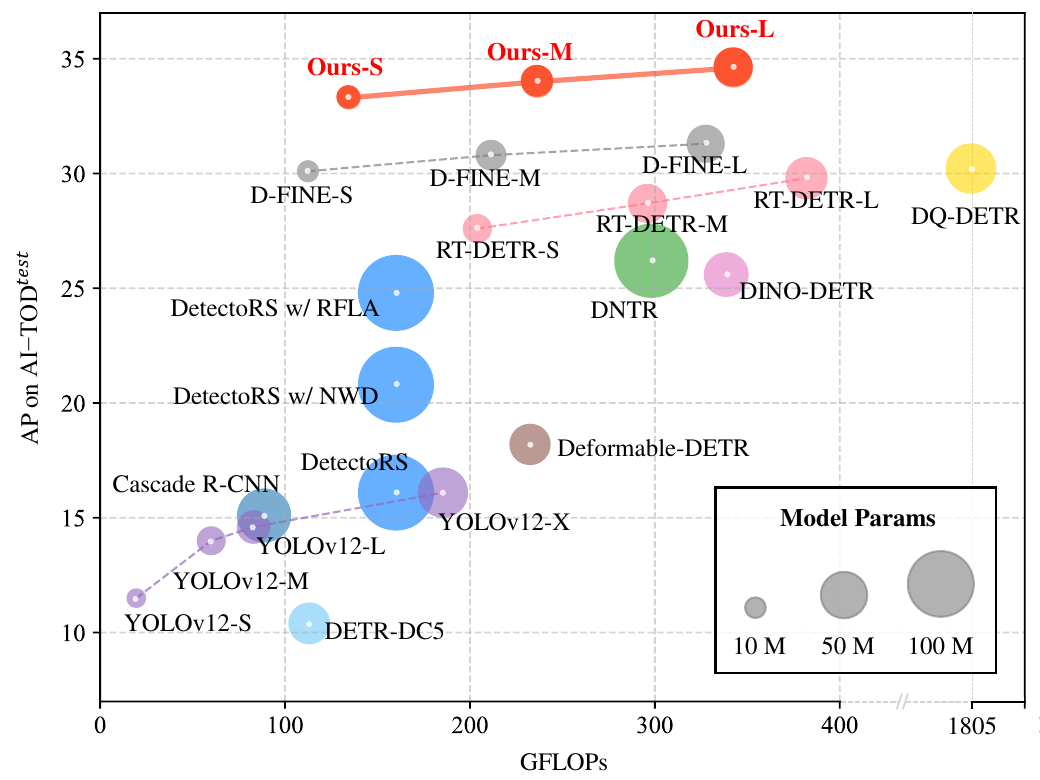}
  \caption{Comparisons with other detectors on the AI-TOD-V2 \textit{test} set. More statistics is on Table \ref{tab:main}.}
  \label{fig:teaser}
\end{figure}

\input{papers/1_Introduction}

\input{papers/2_Related_Work}

\input{papers/3_Method}

\input{papers/4_Experiment}

\input{papers/5_Conclusion}

\section*{Acknowledgements}
This work was in part supported by the National Natural Science Foundation of China under grants 62472399, 62032006 and 62021001.

\bibliographystyle{ACM-Reference-Format}
\bibliography{main.bbl}

\input{appendix/appendix.tex}

\end{document}

%% file: papers/1_Introduction.tex
\section{Introduction}
Object detection is a core task in computer vision, supporting applications like autonomous driving and robotic navigation. Recent progress has been largely driven by CNN-based methods \cite{dai2016r, redmon2016you, he2017mask, lin2017feature, liu2016ssd, girshick2015fast, ren2016faster, wang2023gold, tang2024hic, 10658314}. More recently, Detection Transformers (DETR) introduced Transformer-based end-to-end detection \cite{carion2020end}. Subsequent variants \cite{zhu2020deformable, zhang2022dino, peng2024d, chen2024lw, liu2022dab, meng2021conditional, huang2024deim, zhao2024detrs} have further enhanced performance and speed. However, these advancements primarily target generic object detection, leaving critical challenges unresolved for detecting tiny objects—particularly in aerial imagery from drones or satellites.

Tiny object detection, which focuses on localizing and classifying objects occupying only a few pixels, is critical for tasks like remote sensing, drone monitoring, and autonomous navigation. Its challenge stems from the fragile and sparse feature representation of tiny objects. These objects heavily rely on low-level spatial details, which are often lost in deeper layers of feature hierarchies. Maintaining high-resolution feature maps help preserve such information but comes at the cost of increased computational complexity and memory consumption. Deformable DETR \cite{zhu2020deformable} alleviates some of this burden through sparse deformable attention, yet still suffers from high inference latency due to its broad multi-scale attention. RT-DETR \cite{zhao2024detrs} improves inference speed by decoupling intra-scale and cross-scale interactions but relies heavily on deep, low-resolution features, resulting in degraded performance on small-scale targets. These issues highlight the pressing need for detection frameworks that can balance fine-grained feature preservation with computational efficiency.

Beyond feature representation, query allocation poses another challenge for tiny object detection, especially in aerial images with dense and complex object distributions \cite{wang2021tiny, zhu2021detection}. Existing DETR-like methods \cite{carion2020end, zhu2020deformable, liu2022dab, li2022dn, zhang2022dino, zhao2024detrs} use a fixed number of queries ($K=100$ in DETR, $K=300$ in Deformable-DETR), which simplifies implementation but reduces recall in dense scenes and wastes resources in sparse ones. Some aerial images \cite{wang2021tiny} contain over 1,500 tiny objects, far exceeding typical query capacities. DDQ-DETR \cite{zhang2023dense} increases query density (using $K=900$) and applies class-agnostic NMS with manually set IoU thresholds to filter redundant predictions. However, its rigid query count and fixed NMS thresholds introduce low recall in dense scenes and remain insensitive to variations in instance density. Meanwhile, DQ-DETR \cite{huang2024dq} introduces dynamic query adjustment through a categorical counting module that estimates a classified number for query allocation. Although promising, its counting head relies on manually tuned classification hyperparameters across different datasets, and module design leads to substantial computational overhead. These challenges underscore the need for an adaptive query mechanism that dynamically aligns query density with instance distributions while eliminating manual tuning and preserving efficiency. To this end, tiny object detection remains challenging due to the fragile nature of small object features and the inefficiency of fixed-query mechanisms in handling diverse instance densities. 

\begin{figure}[t]
    \centering
    \includegraphics[width=\linewidth]{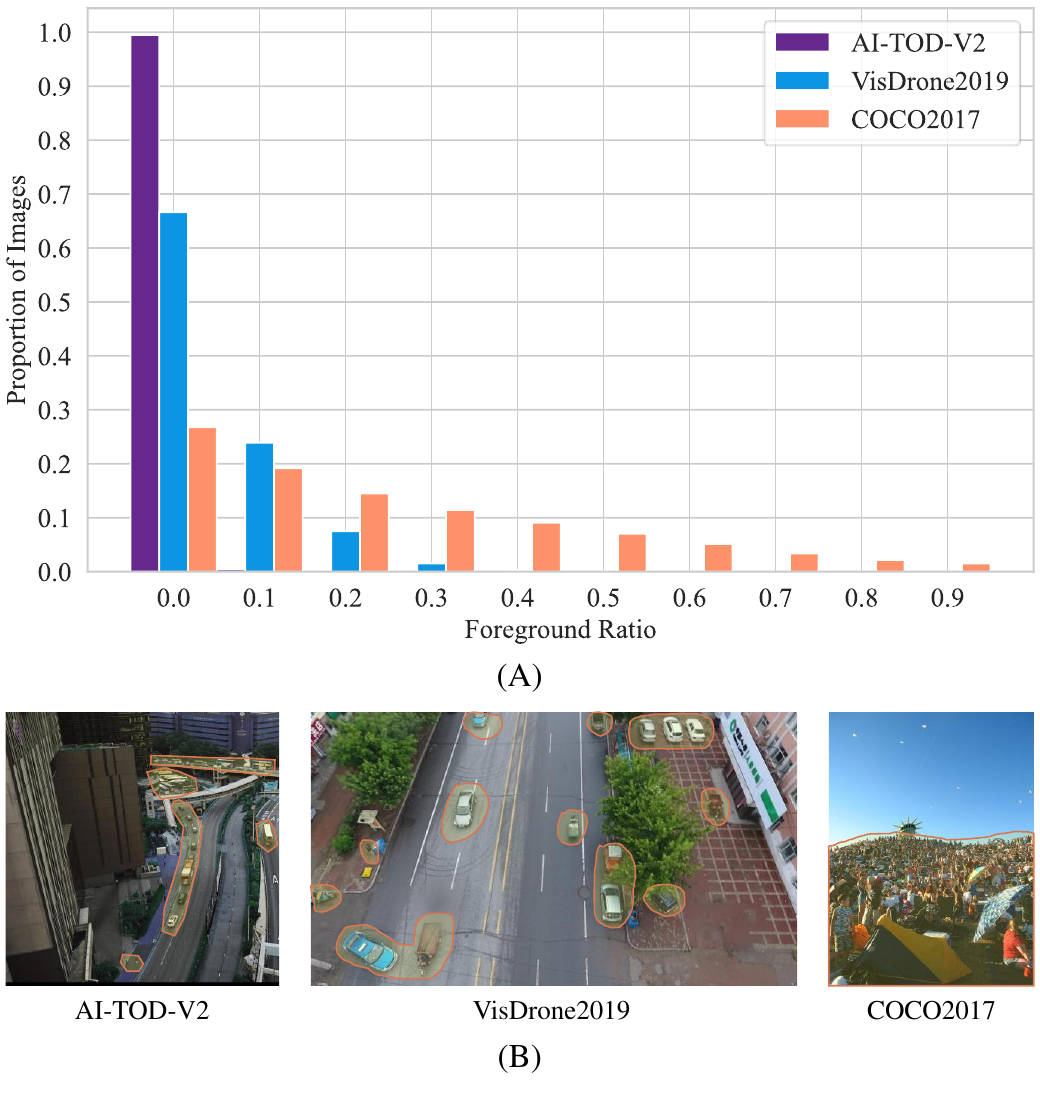}
    \caption{(A) Comparison of foreground proportions across different datasets; and (B) Visualization of foreground regions (highlighted in yellow) in samples from these datasets.}
    \label{fig:foreground}
\end{figure}

In this paper, we introduce \textbf{Dome-DETR}, a novel and efficient end-to-end object detection framework which improves small and tiny object detection. First, in satellite, drone, and natural imagery, the foreground typically occupies only a small portion of the frame (see Figure \ref{fig:foreground}), leading to excessive computation on less informative background regions. This highlights the potential for acceleration by emphasizing foreground areas. Moreover, shallow CNN features inherently capture rich spatial density cues. To leverage this, we design a lightweight \textbf{(A) Density-Focal Extractor (DeFE)} that produces density response heatmaps closely aligned with ground-truth annotations. These heatmaps enhance multi-scale encoder features and guide progressive query initialization in the decoder. Second, while deep features provide scene-level context, instance-level details in shallow features are essential for detecting small objects—but come with high attention costs. To mitigate this, we use DeFE-generated maps to suppress irrelevant regions via \textbf{(B) Masked Window Attention Sparsification (MWAS)}, focusing attention on meaningful windows. Finally, fixed-query mechanisms in DETR struggle with large variations in object counts. We address this with \textbf{(C) Progressive Adaptive Query Initialization (PAQI)}, which adaptively allocates queries by decoding density maps into dynamic suppression thresholds, removing the need for hand-tuned hyperparameters while improving recall in dense scenes.

Experimental results on the AI-TOD-V2 \cite{wang2021tiny} dataset show that Dome-DETR achieves state-of-the-art performance in tiny object detection, outperforming existing models in both accuracy and efficiency. Specifically, Dome-DETR-M and Dome-DETR-L reach 34.0\% (+3.2 AP) and 34.6\% (+3.3 AP) on the AI-TOD-V2 \textit{test} set, with only 252.6 and 358.7 GFLOPs, respectively. On the VisDrone \cite{zhu2021detection} \textit{val} set, Dome-DETR-L also achieves 39.0\% (+2.5 AP). By effectively tackling core challenges in feature representation and query allocation, Dome-DETR marks a substantial step forward for DETR-based tiny object detection. In summary, our main contributions are:

\begin{itemize}

\item We propose Dome-DETR, a novel DETR-based framework for end-to-end tiny object detection, which efficiently enhances feature utilization and query initialization through a finely-tuned density map, improving both accuracy and efficiency.

\item We introduce the Density-Focal Extractor (DeFE) and Masked Window Attention Sparsification (MWAS) to focus computation on informative regions, thereby enhancing both efficiency and detection accuracy.

\item We present Progressive Adaptive Query Initialization (PAQI) to overcome the limitations of rigid query allocation, which adaptively adjusts the number and placement of object queries based on density estimation.

\item We achieve state-of-the-art performance on the AI-TOD-V2 and VisDrone-DET-2019 datasets while maintaining a low computational cost, outperforming all existing models.

\end{itemize}

%% file: papers/2_Related_Work.tex
\section{Related Work}

\subsection{Small / Tiny Object Detection}
Tiny object detection presents significant challenges due to limited pixel information and complex distributions. Traditional CNN-based detectors, such as Faster R-CNN \cite{girshick2015fast, ren2016faster} and FCOS \cite{tian2019fcos}, struggle with small objects due to inadequate feature representation and the lack of long-range dependency modeling. Early solutions focused on data augmentation (e.g., copy-paste strategies \cite{kisantal2019augmentation}) and specialized loss functions \cite{wang2021normalized, xu2022detecting, xu2022rfla, xu2021dot}, which reformulate Intersection over Union (IoU) to consider absolute and relative object sizes. Recent transformer-based models, such as DETR variants \cite{carion2020end, zhu2020deformable, liu2022dab, li2022dn, zhang2022dino, zhao2024detrs, chen2024lw, peng2024d}, mitigate these issues by eliminating hand-crafted components (e.g., NMS) and leveraging self-attention. DQ-DETR \cite{huang2024dq} introduces dynamic query selection, using density maps to adjust query numbers and positions based on instance density. However, these methods heavily depend on manually designed bounding-box representations or finely tuned hyperparameters, making optimization challenging.

\subsection{Real-Time / End-to-End Object Detectors}
The YOLO series has led real-time object detection through advancements in architecture, data augmentation, and training techniques \cite{redmon2016you, wang2023gold}. Despite its efficiency, YOLO relies on Non-Maximum Suppression (NMS), introducing latency and trade-offs in speed and accuracy. DETR \cite{carion2020end} removes hand-crafted components like NMS and anchors but suffers from high computational costs \cite{zhu2020deformable, liu2022dab, li2022dn, zhang2022dino}, limiting real-time use. Recent models—RT-DETR \cite{zhao2024detrs}, LW-DETR \cite{chen2024lw}, and D-FINE \cite{peng2024d}—optimize DETR for real-time applications. Meanwhile, YOLOv10 \cite{wang2024yolov10} eliminates NMS, marking a shift toward fully end-to-end detection. However, these methods underperform on tiny objects due to insufficient focus on shallow features.

\subsection{UAV-specific detector}
Recent UAV-specific detectors have tackled tiny object detection challenges in aerial imagery. QueryDet \cite{yang2022querydet} and ClusDet \cite{yang2019clustered} used coarse-to-fine pipelines for better localization but suffered from high computational costs. Recent UAV-OD methods \cite{tang2024hic, mittal2020deep, shen2023multiple, liu2021survey} are devoted to lightweighting models or optimizing the processing pipeline for practical use. UAV-DETR \cite{zhang2025uav} integrated multi-scale spatial features with frequency-aware processing, using frequency-focused downsampling and semantic calibration to enhance tiny object detection. However, these methods focus solely on small objects and overlook the detection of extremely tiny targets, which remains a significant challenge in UAV-based vision applications.

%% file: papers/3_Method.tex
\begin{figure*}[t]
    \centering
    \includegraphics[width=1.0\linewidth]{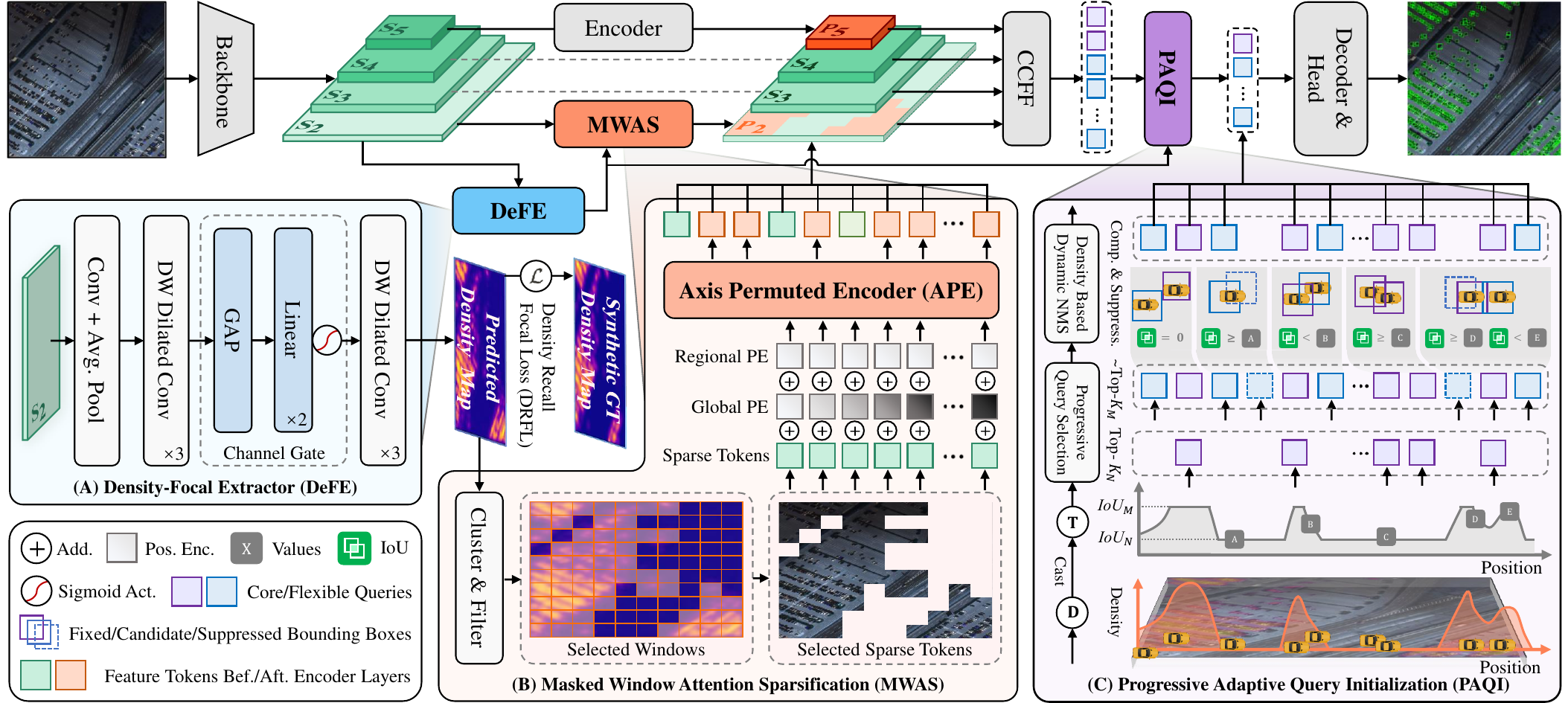}
    \caption{The overall pipeline of our proposed model. The process begins with the backbone network extracting multi-scale features from the input image. Subsequently, (A) the Density-Focal Extractor (DeFE) processes shallow features to generate a predicted density map, which is supervised by our proposed Density Recall Focal Loss (DRFL). These features are then fed into the Encoder, where (B) Masked Window Attention Sparsification (MWAS) selectively processes tokens based on the density map predicted by (A), and the selected tokens are further enhanced by our proposed Axis Permuted Encoder (APE). Concurrently, features are fused and passed to (C) Progressive Adaptive Query Initialization (PAQI), which progressively initializes and refines object queries. Finally, the Decoder and Head layers utilize these refined features and queries to produce the final object detection output.}
    \label{fig:Main}
\end{figure*}

\section{Method}

\subsection{Overview}
As shown in Figure \ref{fig:Main}, our study proposes Dome-DETR, which is built upon D-FINE \cite{peng2024d}. We enhance the model with three components, i.e. (A) Density-Focal Extractor (DeFE) for density prediction, (B) Masked Window Attention Sparsification (MWAS) for efficient shallow feature enhancement, and (C) Progressive Adaptive Query Initialization (PAQI) for dynamic query manipulation.

\subsection{Density-Focal Extractor}
In DETRs, backbone generates spatially-channel encoded feature maps via multi-scale extraction, learned end-to-end to transform basic visual cues into high-level semantics. As a result, foreground object regions exhibit distinctive activations, while backgrounds maintain low responses. Prior studies show that shallow features retain vital spatial cues for small objects \cite{ning2023rethinking, tu2022maxvit}, but they are challenging to exploit due to: (1) High-resolution maps causing prohibitive computation, and (2) Redundant background areas dominating attention, weakening instance-level signals.

To tackle this, we propose the Density-Focal Extractor (DeFE), a lightweight module that explicitly learns instance density distributions to guide efficient feature enhancement and query allocation. As shown in Figure \ref{fig:Main}(A), DeFE processes the shallowest backbone feature map $F_S \in \mathbbm{R}^{H \times W \times C}$ through a cascaded network using depthwise separable convolutions with dilation rates (1, 2, 3) for multi-scale context capture while staying efficient. A lightweight attention block enhances salient areas via channel-wise recalibration. The output passes through a density head with a 3×3 convolution and bilinear upsampling, producing a normalized density heatmap $D_{pred} \in \mathbbm{R}^{H \times W \times 1}$. Thus, DeFE can be formulated as:

\begin{equation}
F_S^{\prime}=f_{\mathrm{DSConv}}(F_S),
\end{equation}
\begin{equation}
F_G=\frac1{HW}\sum_{i=1}^H\sum_{j=1}^WF_S^{\prime}(i,j),
\end{equation}
\begin{equation}
D_{\mathrm{pred}}=\text{Upsample}(\text{Sigmoid}(W_P*F_G)),
\end{equation}
where $f_{\mathrm{DSConv}}(\cdot)$ applies cascaded depthwise separable convolutions with varying dilation rates to extract spatially rich features. $F_G$ is a compact global representation obtained via Global Average Pooling (GAP). $W_{P}$ denotes a convolution layer projecting features into a single-channel density map, followed by a sigmoid function and bilinear upsampling to restore the original resolution and distribution.

The ground-truth density map $D_{gt}$ is generated by convolving Gaussian kernels centered at each object's coordinates, with kernel sizes proportional to their bounding boxes. This encodes both instance locations and relative scales into a continuous supervision signal. To train DeFE, we propose Density Recall Focal Loss (DRFL), which emphasizes accurate density estimation in critical regions:
\begin{equation}
\hspace{-5pt}
\mathcal{L}_{DRFL} = \sum_{i,j}^{H,W}  \left[ \alpha_{i,j}(d_{i,j}^{pred} - d_{i,j}^{gt})^2 + \beta \cdot \mathbbm{1}(d_{i,j}^{pred} < d_{i,j}^{gt}) \cdot d_{i,j}^{gt} \right],
\end{equation}
where $\alpha_{i,j}=\sqrt{d_{i,j}^{gt}}$ adaptively weights positions based on ground-truth density, and $\beta$ penalizes underestimation in dense regions. This ensures balanced learning across sparse and crowded areas while reducing FN in heavy occlusion. Despite high estimation accuracy, the module adds only 0.8M parameters thanks to optimized depthwise convolutions and attention.

\subsection{Masked Window Attention Sparsification}

The high resolution of shallow features is vital for capturing fine details of tiny objects but also incurs significant computational cost. Applying global attention to such dense maps results in high memory usage and latency. To alleviate this, we propose Masked Window Attention Sparsification (MWAS), which selectively focuses computation on key foreground regions while discarding redundant background. As shown in Figure \ref{fig:Main}(B), MWAS comprises two main stages: Foreground Token Pruning and Axis Permuted Encoder (APE).

Using the density map predicted by the DeFE, we first generate a binary mask that retains high-density regions and prunes low-density background. This significantly reduces the tokens involved in attention while preserving essential object details.

\begin{figure}[t]
  \centering
  \includegraphics[width=\linewidth]{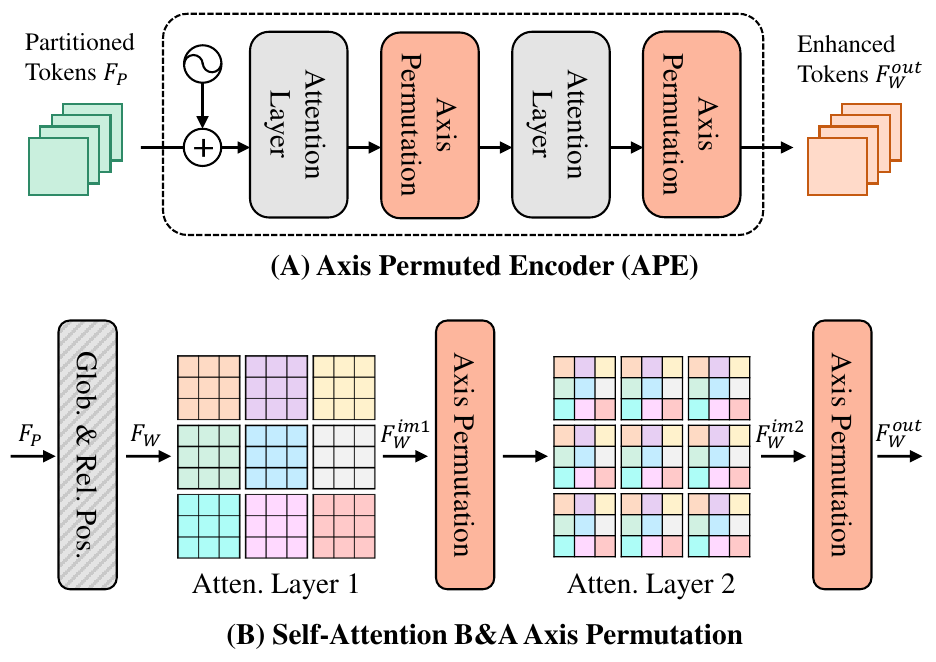}
  \caption{An illustration of the (A) Axis Permuted Encoder (APE) and the (B) Self-Attention Before and After Axis Permutation.}
  \label{fig:ape}
\end{figure}

\textbf{Density-Based Mask Generation.} Given the shallow feature map $F_S$ from the backbone and the density heatmap $D_{\mathrm{pred}}\in\mathbbm{R}^{H\times W\times 1}$ from DeFE, where each element $d_{i,j}^{\mathrm{pred}}$ denotes the estimated object density at position $(i,j)$, we generate a binary mask $M_b \in {0,1}^{H\times W}$ by applying an adaptive threshold $T_b$ to $D_{\mathrm{pred}}$:
\begin{equation}
M_b^{(i,j)} = \mathbbm{1}\left(d_{i,j}^{\mathrm{pred}} > T_b\right),
\end{equation}
where $T_b$ is determined by the minimal adjustment steps $k^*$ required to activate at least one foreground region:  
\begin{equation}
k^* = \min\left[ k \in \mathbbm{N} \,\bigg|\, \sum_{i,j} \mathbbm{1}\left(d_{i,j}^{\mathrm{pred}} > T_{\mathrm{init}} - k \Delta T\right) > 0 \right],
\end{equation}
\begin{equation}
T_b = T_{\mathrm{init}} - k^* \Delta T,
\end{equation}
where $T_{\mathrm{init}}$ denotes the initial threshold and $\Delta T$ is the decrement step. This formulation ensures $T_b$ is the highest threshold satisfying $\sum_{i,j} M_b^{(i,j)} > 0$, preserving meaningful object regions while suppressing background redundancy.

\textbf{Window Partitioning and Background Token Pruning.} The shallow backbone feature map $F_S$ is divided into non-overlapping windows of size $(h,w)$, ensuring structured processing. To determine which windows contain valid foreground information, a window-level mask is computed by applying max pooling over the binary mask $M_{b}$:
\begin{equation} M_W^{(i,j)} = \max_{(p,q) \in W_{i,j}} M_b^{(p, q)}, \end{equation}
where $W_{i,j}$ represents the set of pixels within the $(i,j)$-th window. If $M_W^{(i,j)}=1$, the window is retained for further processing. The selected $k$ windows are then collected as $F_P \in \mathbbm{R}^{k \times h\times w\times C}$.

\input{figures/paqi}

Secondly, attention is efficiently computed within each window, enhancing object details while minimizing redundant global interactions. To improve cross-window communication, we introduce axis permuted attention, enabling both regional focus and long-range dependencies across high-confidence areas.

\input{tables/main}

\textbf{Axis Permuted Encoder for Feature Enhancement.} For each valid window, the corresponding feature map $F_{P}^{(i,j)}$ is augmented with relative and global positional encodings. The Axis Permuted Encoder (APE) then refines features by processing these windows with sequential self-attention, incorporating a spatial permutation mechanism, as shown in Figure \ref{fig:ape}. Given the position-encoded window feature $F_{W,k}^{(i,j)}$, where $k$ is the intermediate state index, the first self-attention is computed as:
\begin{equation}
Q^{(i,j)}_1 = K^{(i,j)}_1 = V^{(i,j)}_1 = F_W^{(i,j)}, \end{equation}
\begin{equation}
F^{(i,j)}_{W, 1} = \text{FFN}(\text{MSA}(Q^{(i,j)}, K^{(i,j)}, V^{(i,j)})), \end{equation}
where $\mathrm{MSA}(\cdot)$ denotes multi-head self-attention, capturing intra-window dependencies. To establish long-range spatial interactions, we permute feature axes and apply a second self-attention pass:
\begin{equation}
Q^{(i,j)}_2 = K^{(i,j)}_2 = V^{(i,j)}_2 = \text{Permute}(F^{(i,j)}_{W, 1}),
\end{equation}
\begin{equation}
F^{(i,j)}_{W, 2} = \text{MSA}(Q^{(i,j)}_2, K^{(i,j)}_2, V^{(i,j)}_2),
\end{equation}
where $\mathrm{Permute}(\cdot)$ rearranges spatial dimensions to propagate information across windows. The refined features are then passed through a feedforward network (FFN) with residual connections:
\begin{equation}
F_{W, \text{out}}^{(i,j)} = \text{FFN}(\text{Permute}(F^{(i,j)}_{W, 2})). \end{equation}

This encoding scheme effectively captures both local and global dependencies while maintaining computational efficiency, allowing MWAS to focus resources on high-confidence object regions.

\subsection{Progressive Adaptive Query Initialization}
Existing DETR variants typically use a fixed number of queries (e.g., 300), selected based on token scores. However, this static allocation fails to adapt to varying object densities, leading to inefficiency in sparse scenes and insufficient queries in dense ones. Prior works have attempted solutions via categorized query counts \cite{huang2024dq} or dense queries \cite{zhang2023dense}, but these lack adaptability to object distribution and require extra hyperparameter tuning, increasing complexity across datasets.

We propose \textbf{Progressive Adaptive Query Initialization (PAQI)}, which dynamically adjusts query density based on scene complexity without manual threshold tuning. Our method is based on the observation that Top-K encoder queries are often redundant in sparse regions and inadequate in dense ones. Many are also initialized in background regions, offering no benefit. To address this, PAQI progressively initializes queries by focusing on high-response regions in the density map while filtering out low-response areas. This eliminates dataset-specific tuning and enables adaptive, scene-aligned query generation.

\textbf{Candidate Query Selection via Classification Scores.} As shown in Algorithm \ref{algo:1}, PAQI first generates anchors based on the image’s spatial distribution. The encoded feature memory is passed through a classification head to obtain objectness scores, reflecting the likelihood of each anchor containing an object. To balance redundancy and coverage in dense areas, the Top-$K_M$ features by score are selected as candidate queries.

\textbf{Query Splitting and Density-Aware Sampling.} To improve robustness under varying densities, the selected queries are split into: (1) a core set of the first $K_N$ queries as a strong baseline, and (2) a flexible set of the remaining $K_M - K_N$ for refinement. Density-aware filtering is applied using a high-response mask from DeFE, discarding low-response queries to avoid oversampling sparse regions while ensuring dense areas receive sufficient attention.

\textbf{Bounding Box Generation and Dynamic NMS.} To further adapt query count to density, refined queries predict bounding boxes via a regression head, followed by density-aware dynamic Non-Maximum Suppression (NMS).

Traditional NMS uses a fixed IoU threshold, which may cause excessive suppression in dense regions or too many anchors in sparse areas. To address this, we dynamically adjust the IoU threshold:
\begin{equation}
\mathbf{T} = IoU_N + \mathbf{D} \times (IoU_M - IoU_N),
\end{equation}
where $\mathbf{D}$ is the predicted density of the final mask-filtered tokens, and the IoU threshold adapts to it. This avoids over-suppression in dense areas while reducing queries in sparse regions. Specifically, a higher IoU threshold is used in dense regions to prevent excessive suppression, and a stricter threshold in sparse regions to improve efficiency, enhancing both detection performance and speed.

%% file: figures/paqi.tex
\begin{algorithm}[t]
\caption{Progressive Query Initialization for Object Detection}
\label{algo:1}
\begin{algorithmic}[1]
\STATE \textbf{Input:} Memory $\mathbf{M} \in \mathbb{R}^{B \times N \times C}$, Window Mask $\mathbf{W}$, Feature Map $\mathbf{F}$, Density Map $\mathbf{D}$
\STATE \textbf{Output:} Query Features $\mathbf{Q}$, Bounding Boxes $\mathbf{B}$, Classification Scores $\mathbf{S}_{cls}$
\STATE \textbf{Hyperparameters:} Max Query Nums $K_M$, Min Query Nums $K_N$, Max IoU Thres. $IoU_{M}$, Min IoU Thres. $IoU_{N}$

\STATE \textcolor{gray}{\# \textit{Generate initial anchors and valid mask}}
\STATE \( \mathbf{A} \gets \text{GenerateAnchors}(\mathbf{S}) \)
\STATE \( \mathbf{M}_{out}, \mathbf{S}_{enc} \gets \text{ClsHead}(\mathbf{M}) \) 

\STATE \textcolor{gray}{\# \textit{Select Top-$K_M$ features based on scores}}
\STATE \( \mathbf{M}_{T}, \mathbf{S}_{T}, \mathbf{A}_{T} \gets \text{SelectTopK}( \mathbf{M}_{out}, \mathbf{S}_{enc}, \mathbf{A}, K_M) \)

\STATE \textcolor{gray}{\# \textit{Split into core and flexible queries}}
\STATE \( \mathbf{M}_1, \mathbf{S}_1, \mathbf{A}_1 \gets \mathbf{M}_{T}[:K_{N}], \mathbf{S}_{T}[:K_{N}], \mathbf{A}_{T}[:K_{N}] \)
\STATE \( \mathbf{M}_2, \mathbf{S}_2, \mathbf{A}_2 \gets \mathbf{M}_{T}[K_{N}:K_{M}] , \mathbf{S}_{T}[K_{N}:K_{M}], \mathbf{A}_{T}[K_{N}:K_{M}] \)

\STATE \textcolor{gray}{\# \textit{Filter flexible queries based on high density response mask}}
\STATE \(\mathbf{A}_2 \gets \mathbf{A}_2[\mathbf{W}[\mathbf{A}_2] > 0] \)

\STATE \( \mathbf{M}_{final}, \mathbf{A}_{final}, \mathbf{S}_{final} \gets \text{Concat}\{(\mathbf{M}_1, \mathbf{M}_2), (\mathbf{A}_1, \mathbf{A}_2),
(\mathbf{S}_1, \mathbf{S}_2)\}\)

\STATE \textcolor{gray}{\# \textit{Generate bounding boxes and apply dynamic NMS}}
\STATE \( \mathbf{B}_{raw} \gets \text{BboxHead}(\mathbf{M}_{final}, \mathbf{A}_{final}) \)
\STATE \( \mathbf{T} \gets  IoU_{N} + \mathbf{D} \times (IoU_{M} - IoU_{N})\)
\STATE \( \mathbf{B}_{final} \gets \text{DynamicNMS}(\mathbf{B}_{raw}, \mathbf{S}_{final}, \mathbf{T}) \)

\STATE \textbf{Return} $\mathbf{M}_{final}, \mathbf{B}_{final}, \mathbf{S}_{final}$
\end{algorithmic}
\end{algorithm}

%% file: tables/main.tex
\begin{table*}[ht]
    \tabcolsep=0.28cm
    \centering
    \small
    \renewcommand{\arraystretch}{1.2}
    \caption{Experiments on AI-TODV2. All models are trained on the \textit{trainval} split and evaluated on the \textit{test} split with 800 $\times$ 800 input resolution. $\dagger$ notes a re-implementation of the results. * denotes re-implementation with 4 feature map layers for fair comparison. $^{\star}$ denotes an average value across all data in the split.}
    \begin{tabular}{lcccccccccc}
        \toprule
        \textbf{Method} & \textbf{\#Params.} & \textbf{GFLOPs} & \textbf{AP} & \textbf{AP$_{50}$} & \textbf{AP$_{75}$} & \textbf{AP$_{vt}$} & \textbf{AP$_{t}$} & \textbf{AP$_{s}$} & \textbf{AP$_{m}$} \\
        \midrule
        \arrayrulecolor{black} \specialrule{0.5pt}{0pt}{0pt} 
        \rowcolor[HTML]{F2ECDE} \multicolumn{10}{l}{\textit{Non-end-to-end Object Detectors}} \\
        Cascade R-CNN \scriptsize \textcolor{gray}{[CVPR 2018]} \normalsize \cite{cai2018cascade} & 68.9M & 88.7    & 15.1  & 34.2  & 11.2  & 0.1  & 11.5  & 26.7  & 38.5  \\
        DetectoRS \scriptsize \textcolor{gray}{[Arxiv 2020]} \normalsize \cite{qiao2021detectors} & 134.8M    & $\approx160$ & 16.1  & 35.5  & 12.5  & 0.1  & 12.6  & 28.3  & 40.0  \\
        DetectoRS w/ NWD \scriptsize \textcolor{gray}{[ISPRS 2022]} \normalsize \cite{wang2021normalized}& -- & --    & 20.8  & 49.3  & 14.3  & 6.4  & 19.7  & 29.6  & 38.3  \\
        DetectoRS w/ RFLA \scriptsize \textcolor{gray}{[ECCV 2022]} \normalsize \cite{xu2022rfla} & --   & --    & 24.8  & 55.2  & 18.5  & 9.3  & 24.8  & 30.3  & 38.2  \\
        DNTR \scriptsize \textcolor{gray}{[TGRS 2024]} \normalsize \cite{liu2024denoising} & 128.4M & 298.1 & 26.2  & 56.7  & 20.2  & 12.8  & 26.4  & 31.0  & 37.0  \\
        \midrule
        \arrayrulecolor{black} \specialrule{0.5pt}{0pt}{0pt} 
        \rowcolor[HTML]{F2ECDE} \multicolumn{10}{l}{\textit{End-to-end Object Detectors}} \\
        DETR-DC5 \scriptsize \textcolor{gray}{[ECCV 2020]} \normalsize \cite{carion2020end}& 41M   & 113.1    & 10.4  & 32.5  & 3.9   & 3.6   & 9.3   & 13.2  & 24.6  \\
        YOLOv12-S$\dagger$ \scriptsize \textcolor{gray}{[Arxiv 2025]} \normalsize \cite{tian2025yolov12}& 9.1M  & 19.6  & 11.5  & 25.4  & 9.0  & 2.7  & 11.8  & 18.0  & 23.7  \\
        YOLOv12-M$\dagger$ \scriptsize \textcolor{gray}{[Arxiv 2025]} \normalsize \cite{tian2025yolov12}& 19.6M & 60.1  & 14.0  & 30.7  & 11.1  & 3.6  & 14.6  & 20.0  & 26.3  \\
        YOLOv12-L$\dagger$ \scriptsize \textcolor{gray}{[Arxiv 2025]} \normalsize \cite{tian2025yolov12}& 26.5M & 83.0  & 14.6  & 31.0  & 12.0  & 4.1  & 15.3  & 20.3  & 27.6  \\
        YOLOv12-X$\dagger$ \scriptsize \textcolor{gray}{[Arxiv 2025]} \normalsize \cite{tian2025yolov12}& 59.3M & 185.4 & 16.1  & 33.5  & 13.4  & 5.4  & 17.0  & 21.4  & 28.5  \\
        Deformable-DETR* \scriptsize \textcolor{gray}{[ICLR 2021]} \normalsize \cite{zhu2020deformable}& 40M & 232.5    & 18.2  & 59.7  & 10.2  & 5.8   & 16.8  & 25.2  & 34.0  \\
        DINO-DETR* \scriptsize \textcolor{gray}{[ICLR 2023]} \normalsize \cite{zhang2022dino}& 47M  & 338.6    & 25.6  & 61.3  & 17.4  & 12.7  & 25.3  & 31.7  & 39.4  \\
        DQ-DETR* \scriptsize \textcolor{gray}{[ECCV 2024]} \normalsize \cite{huang2024dq}& 58.7M & $1805.4^{\star}$ & 30.2  & 68.6  & 22.3  & 15.3  & 30.5  & 36.5  & 44.6  \\
        D-FINE-S* \scriptsize \textcolor{gray}{[ICLR 2024]} \normalsize \cite{peng2024d}& 11.6M & 112.4 & 30.1  & 64.1  & 25.5  & 14.0  & 30.4  & 36.0  & 45.4  \\
        D-FINE-M* \scriptsize \textcolor{gray}{[ICLR 2024]} \normalsize \cite{peng2024d}& 22.3M & 211.4 & 30.8  & 65.4  & 25.3  & 14.3  & 30.1  & 37.6  & 46.8  \\
        D-FINE-L* \scriptsize \textcolor{gray}{[ICLR 2024]} \normalsize \cite{peng2024d}& 34M   & 327.5 & 31.3    & 66.7    & 24.9    & 15.3    & 31.0    & 38.4    & 46.6    \\
        \rowcolor[HTML]{F5F5F5} \textbf{Dome-DETR-S} (Ours) & 13.2M & $154.2^{\star}$    & 33.3 \color{purple}(+3.2)    & 67.5    & 28.9    & 17.8    & 33.0    & 38.4    & 46.4    \\
        \rowcolor[HTML]{EBEBEB} \textbf{Dome-DETR-M} (Ours) & 23.9M & $252.6^{\star}$    & 34.0 \color{purple}(+3.2) & 68.4  & 29.9  & 18.4  & 34.3  & 39.0  & 46.9  \\
        \rowcolor[HTML]{E1E1E1} \textbf{Dome-DETR-L} (Ours) & 36.0M & $358.7^{\star}$ & \textbf{34.6} \color{purple}\textbf{(+3.3)} & \textbf{69.2}  & \textbf{32.0}  & \textbf{19.0}  & \textbf{35.6}  & \textbf{39.8}  & \textbf{47.3}  \\
        \bottomrule
    \end{tabular}
    \label{tab:main}
\end{table*}

%% file: papers/4_Experiment.tex
\section{Experiment}
\label{experiment}

\subsection{Datasets}

We conduct experiments on two aerial datasets: AI-TOD-V2 and VisDrone, both of which primarily contain tiny and small objects.

\textbf{AI-TOD-V2.} AI-TOD-V2 \cite{xu2022detecting} consists of 28,036 aerial images with a total of 752,745 annotated object instances. The dataset is split into three subsets: 11,214 images for training, 2,804 for validation, and 14,018 for testing. The dataset is characterized by extremely small object sizes, with an average object size of only 12.7 pixels. Notably, 86\% of the objects are smaller than 16 pixels, and even the largest object does not exceed 64 pixels. Additionally, the number of objects per image varies significantly, ranging from 1 to 2,667, with an average of 24.64 objects per image and a standard deviation of 63.94.

\textbf{VisDrone.} VisDrone \cite{zhu2021detection} comprises 14,018 drone-captured images, with 6,471 images in the training set, 548 in the validation set, and 3,190 in the test set. The dataset spans 10 categories. It features a diverse range of objects, including pedestrians, vehicles, and bicycles, and exhibits variations in object density, from sparse to highly crowded scenes. The average number of objects per image is 40.7, with a standard deviation of 46.41.

\input{tables/main_vis}

\begin{figure*}[t]
    \centering
    \includegraphics[width=1\linewidth]{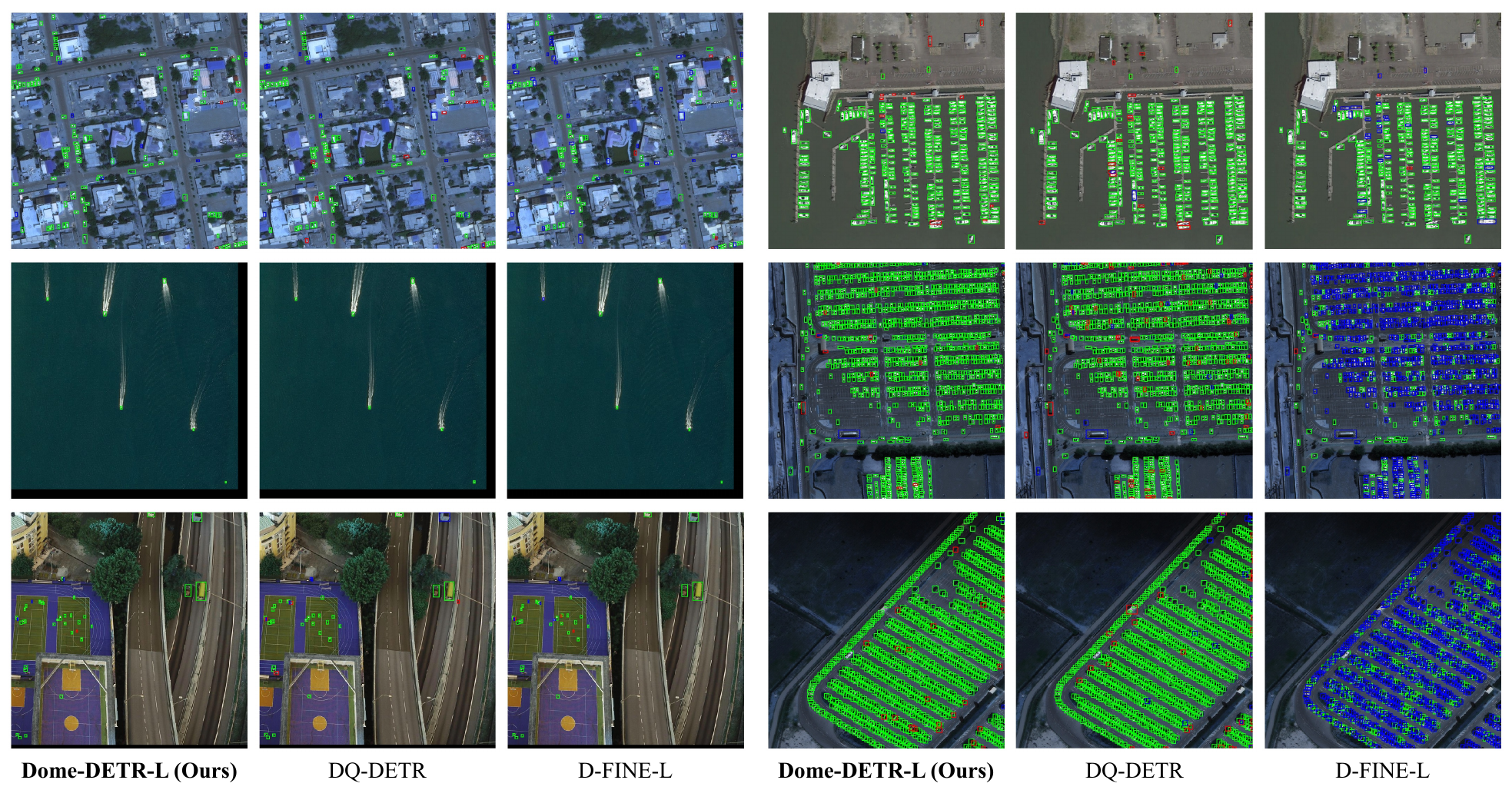}
    \caption{Visualization of comparison between our proposed method and other methods under different scenes on AI-TOD-V2 \textit{test} split. The green, red, and blue boxes represent TP, FP, and FN, respectively.}
    \label{fig:visual:main}
\end{figure*}

\subsection{Evaluation Metric}
To assess the performance of our proposed method, we use the Average Precision (AP) metric with a maximum detection limit of 1,500 objects. Specifically, AP is calculated as the average AP value over the range of IoU thresholds from 0.50 to 0.95, with a step size of 0.05. Furthermore, we employ scale-specific AP evaluations in AI-TOD-V2\cite{faster-coco-eval-aitod, faster-coco-eval}, including $AP_{vt}$, $AP_t$, $AP_s$, and $AP_m$, which correspond to very tiny, tiny, small, and medium-sized object evaluations, as defined in \cite{wang2021tiny}.

\input{tables/main_ablation}

\input{tables/main_hyperparameter}

\subsection{Implementation Details}
Based on the D-FINE structure \cite{peng2024d}, we employ a 1-layer transformer encoder, a deformable transformer decoder, and HGNetv2 as our CNN backbone \cite{zhao2024detrs}. Our model is trained on $8\times$ NVIDIA 4090 GPUs. Following D-FINE \cite{peng2024d}, we apply the same random crop and scale augmentation strategies. Additionally, we apply mixup \cite{zhang2017mixup} and Mosaic \cite{bochkovskiy2020yolov4} techniques for VisDrone training, with Mosaic set to a probability of 1 and mixup set to a probability of 0.2.

We provide three versions of our model: \textbf{Small}, \textbf{Medium}, and \textbf{Large}, with increasing parameter size and computational complexity. Detailed hyperparameter configurations for different Dome-DETR models can be found in Appendix D.

\subsection{Main Results}

\textbf{AI-TOD-V2}. Table 2 presents our main results on the AI-TOD-V2 \cite{xu2022detecting} test split. We compare Dome-DETR with strong baselines, including both non-end-to-end and end-to-end detectors. Except for YOLOv12, all non-end-to-end detectors use ResNet50 with FPN \cite{he2016deep}. We re-implemented several DETR-like models on AI-TOD-V2; all except DETR use 4-scale feature maps from backbone stages 1–4. Results in Table \ref{tab:main} show our Dome-DETR-M and Dome-DETR-L achieve the best AP scores of 34.0 and 34.6, surpassing other state-of-the-art methods. Dome-DETR-L improves over the baseline by +3.7\%, +4.6\%, +1.4\%, and +0.7\% in $AP_{vt}$, $AP_{t}$, $AP_{s}$, and $AP_{m}$, with greater gains in $AP_{vt}$ and $AP_{t}$. Our model outperforms advanced DETR-like methods on AI-TOD-V2. LRP evaluation results are in Appendix A.

\textbf{VisDrone.} Table \ref{tab:main_vis} presents results of VisDrone \cite{zhu2021detection} on the validation split. Our Dome-DETR-L achieves the best AP (39.0\%), surpassing D-FINE-L by +2.5\% AP. Similarly, Dome-DETR-M outperforms D-FINE-M by +2.5\% AP, while Dome-DETR-S exceeds D-FINE-S by +2.3\% AP. Compared to DQ-DETR, Dome-DETR-L improves AP, AP$_{50}$, and AP$_{75}$ by +3.8\%, +6.2\%, and +5.2\%, respectively.

\subsection{Efficiency Analysis}
Despite its enhanced capabilities, Dome-DETR maintains competitive efficiency. As shown in Table \ref{tab:main}, Dome-DETR achieves superior accuracy while maintaining competitive efficiency. Compared to D-FINE-L (34M params, 327.5 GFLOPs), Dome-DETR-L (36M params, 358.7 GFLOPs on average) improves AP by 2.5 points with only a 5.9\% parameter increase and only 31.2 higher average GFLOPs. Meanwhile, Dome-DETR-M and Dome-DETR-S also achieves better performance while maintaining efficiency, demonstrating the effectiveness of our density-guided sparsification. More analysis could be found in Appendix B.

\subsection{Ablation Study}
Density-Focal Extractor (DeFE), Masked Window Attention Sparsification (MWAS) and Progressive Adaptive Query Initialization (PAQI) are the newly proposed contributions. We conduct a series of ablation studies to verify the effectiveness of each component proposed in this paper, with D-FINE-S chosen as the comparison DETR-like baseline.

\textbf{Main ablation experiment.} Table \ref{tab:main_ablation} shows the performance of our contributions separately on AI-TOD-V2. Incorporating each proposed component leads to a notable improvement in performance. Specifically, adding DeFE alone enhances AP from 30.1 to 31.2, while integrating PAQI further boosts it to 32.1. The full model, which combines all three components, achieves the best performance, reaching an AP of 33.3. This demonstrates the complementary benefits of DeFE, MWAS, and PAQI in refining tiny object detection capabilities.

\textbf{Hyperparameter Sensitivity Analysis.} Table \ref{tab:ablation_hyperparameter} highlights the impact of key hyperparameters on Dome-DETR-S performance. For IoU$_N$ and IoU$_M$, the best AP (33.3) is achieved at (0.4, 0.9), as a moderate negative threshold prevents excessive suppression of true positives, while a stricter positive threshold ensures high-quality detections. Lowering IoU$_M$ to 0.7 slightly reduces AP due to increased ambiguity in positive assignments, whereas increasing IoU$_N$ to 0.6 marginally drops AP by over-filtering hard positives. For the query initialization threshold $T_{\mathrm{init}}$, 0.05 provides the best balance, achieving AP 32.6. Raising it to 0.1 slightly decreases AP (32.9) due to excessive pruning, whereas lowering it to 0.03 significantly degrades AP (32.3) by introducing unnecessary background noise. For window size $H/W$, $H/W = 10$ offers the best trade-off. A smaller $H/W$ enhances very tiny objects but lowers overall AP, while a larger $H/W$ benefits medium-sized objects at the cost of smaller object performance. This suggests an intermediate window resolution balances fine detail and global context.

\subsection{Visualization Analysis}
Figure \ref{fig:visual:main} presents a comparative analysis of our method against other models across different detection scenarios on the AI-TOD-V2 \textit{test} split. By incorporating density-oriented feature-query manipulation, Dome-DETR achieves performance improvements in both dense and sparse scenes. Additional visualizations of the intermediate processing steps are provided in Appendix C.

%% file: tables/main_vis.tex
\begin{table}[t]
    \tabcolsep=0.09cm
    \centering
    \small
    \renewcommand{\arraystretch}{1.2}
    \caption{Experiments on VisDrone. All models are trained on the \textit{train} split and evaluated on the \textit{val} split with 800 $\times$ 800 input resolution. $\dagger$ notes a re-implementation of the results. $^{\star}$ denotes an average value across all data in the split.}
    \begin{tabular}{lccccc}
        \toprule
        \textbf{Method} & \textbf{\#Params.} & \textbf{GFLOPs} & \textbf{AP} & \textbf{AP$_{50}$} & \textbf{AP$_{75}$} \\
        \midrule
        \arrayrulecolor{black} \specialrule{0.5pt}{0pt}{0pt}
        \rowcolor[HTML]{F2ECDE} \multicolumn{6}{l}{\textit{Non-end-to-end Object Detectors}} \\
        Cascade R-CNN \cite{cai2018cascade}& 68.9M & 236.4 & 22.6 & 38.8 & 23.2 \\
        DetectoRS w/RFLA \cite{xu2022rfla}& 134.8M & 160.0 & 27.4 & 45.3 & 23.9 \\
        CEASC$\dagger$ \cite{du2023adaptive}& $\approx90$M & $150.2^{\star}$ & 28.7 & 50.7 & 24.7 \\
        DNTR \cite{liu2024denoising}& 128.37M & 373.4 & 33.1 & 53.8 & 34.8 \\
        \midrule
        \arrayrulecolor{black} \specialrule{0.5pt}{0pt}{0pt}
        \rowcolor[HTML]{F2ECDE} \multicolumn{6}{l}{\textit{End-to-end Object Detectors}} \\
        YOLOv12-L$\dagger$ \cite{tian2025yolov12}& 26.5M & 82.4 & 25.9 & 39.1 & 28.1 \\
        YOLOv12-X$\dagger$ \cite{tian2025yolov12}& 59.3M & 184.6 & 27.7 & 42.1 & 29.9 \\
        UAV-DETR$\dagger$ \cite{zhang2025uav}& 45.5M & 247.1 & 32.4 & 51.9 & 33.6 \\
        DQ-DETR$\dagger$ \cite{huang2024dq}& 58.7M & $1782.4^{\star}$ & 35.2 & 54.9 & 35.6 \\
        D-FINE-S$\dagger$ \cite{peng2024d}& 11.6M & 112.4 & 31.2 & 53.8 & 36.1 \\
        D-FINE-M$\dagger$ \cite{peng2024d}& 22.3M & 211.4 & 33.6 & 56.9 & 37.6 \\
        D-FINE-L$\dagger$ \cite{peng2024d}& 34.4M & 327.5 & 36.5 & 58.0 & 38.2 \\
        \rowcolor[HTML]{F5F5F5} \textbf{Dome-DETR-S} (Ours)& 13.2M & $176.5^{\star}$ & 33.5 \color{purple}(+2.3)& 56.6 & 37.8 \\
        \rowcolor[HTML]{EBEBEB} \textbf{Dome-DETR-M} (Ours)& 23.9M & $284.6^{\star}$ & 36.1 \color{purple}(+2.5) & 59.8 & 39.4 \\
        \rowcolor[HTML]{E1E1E1} \textbf{Dome-DETR-L} (Ours)& 36.0M & $376.4^{\star}$ & \textbf{39.0} \color{purple}\textbf{(+2.5)} & \textbf{61.1} & \textbf{40.8} \\
        \bottomrule
    \end{tabular}
    \label{tab:main_vis}
\end{table}

%% file: tables/main_ablation.tex
\begin{table}[t]
    \centering
    \tabcolsep=0.205cm
    \renewcommand{\arraystretch}{1.2}
    \caption{Overall ablation for our proposed Dome-DETR on AI-TOD-V2 \textit{test} split.}
    \small
    \begin{tabular}{ccc|cccccc}
        \toprule
        \textbf{DeFE} & \textbf{MS-WAS} & \textbf{PAQI} & \textbf{AP} & \textbf{AP$_{vt}$} & \textbf{AP$_{t}$} & \textbf{AP$_{s}$} & \textbf{AP$_{m}$} \\
        \midrule
        \ding{55} & \ding{55} & \ding{55} & 30.1 & 14.0 & 30.4 & 36.0 & 45.4 \\
        \ding{51} & \ding{51} & \ding{55} & 31.2 & 17.4 & 31.9 & 36.6 & 45.4 \\
        \ding{51} & \ding{55} & \ding{51} & 32.1 & 16.9 & 32.5 & 37.6 & 45.9 \\
        \rowcolor[HTML]{F5F5F5} \ding{51} & \ding{51} & \ding{51} & \textbf{33.3} & \textbf{17.8} & \textbf{33.0} & \textbf{38.4} & \textbf{46.4} \\
        \bottomrule
    \end{tabular}
    \label{tab:main_ablation}
\end{table}

%% file: tables/main_hyperparameter.tex
\begin{table}[t]
    \caption{Hyperparameter ablation studies of Dome-DETR-S on AI-TOD-V2 \textit{test} split.}
    \centering
    \renewcommand{\arraystretch}{1.2}
    \small
    \tabcolsep=0.402cm

    \begin{minipage}[t]{\linewidth}
        \centering
        \hspace*{-0.385cm}
        \begin{minipage}{0.7\linewidth}
            \centering
            \begin{tabular}{cc|cc}
                \toprule
                \textbf{IoU$_{N}$} & \textbf{IoU$_{M}$} & \textbf{AP}  & \textbf{AR}  \\
                \midrule
                \rowcolor[HTML]{F5F5F5}\textbf{0.4} & \textbf{0.9} & \textbf{33.3} & \textbf{48.6} \\
                0.4 & 0.7 & 32.5 & 47.2 \\
                0.6 & 0.9 & 32.7 & 48.6 \\
                \bottomrule
            \end{tabular}
        \end{minipage}%
        \begin{minipage}{0.3\linewidth}
            \centering
            \begin{tabular}{c|c}
                \toprule
                \textbf{T$_{init}$} & \textbf{AP} \\
                \midrule
                0.1 & 32.9 \\
                \rowcolor[HTML]{F5F5F5} \textbf{0.05} & \textbf{33.3} \\
                0.03 & 32.3 \\
                \bottomrule
            \end{tabular}
        \end{minipage}
    \end{minipage}

    \vspace{0.15cm}

    \begin{minipage}{\linewidth}
        \centering
        \tabcolsep=0.43cm
        \begin{tabular}{l|ccccc}
            \toprule
            \textbf{H/W} & \textbf{AP} & \textbf{AP$_{vt}$} & \textbf{AP$_{t}$} & \textbf{AP$_{s}$} & \textbf{AP$_{m}$} \\
            \midrule
            5 & 31.8 & \textbf{17.9} & 33.0 & 36.2 & 44.7 \\
            \rowcolor[HTML]{F5F5F5} \textbf{10} & \textbf{33.3} & 17.8 & \textbf{33.0} & \textbf{38.4} & 46.4 \\
            20 & 32.4 & 17.3 & 32.7 & 38.4 & \textbf{46.8} \\
            \bottomrule
        \end{tabular}
    \end{minipage}

    \label{tab:ablation_hyperparameter}
\end{table}

%% file: papers/5_Conclusion.tex
\section{Conclusion}
In this paper, we present Dome-DETR, a novel end-to-end object detection framework tailored for tiny object detection. Motivated by the unique challenges of small object detection including inefficient feature utilization and imbalanced query allocation, we introduce Density-Focal Extractor (DeFE), Masked Window Attention Sparsification (MWAS), and Progressive Adaptive Query Initialization (PAQI) to effectively addresses these issues, therefore achieving state-of-the-art results on the AI-TOD-V2 and VisDrone datasets in both accuracy and efficiency. Future work may explore more advanced architectures or bounding box representations to further enhance DETR-like models for small object detection. We hope Dome-DETR inspires continued progress in this area—unfolding new possibilities beneath the vast celestial dome.

%% file: appendix/appendix.tex














\appendix

\makeatletter
\twocolumn[{
  \begin{center}
    \Huge \bfseries \@titlefont \textbf{Dome-DETR: DETR with Density-Oriented Feature-Query Manipulation for Efficient Tiny Object Detection}\\
  \end{center}
  \begin{center}
    \Huge \textit{Appendix}\\[1em]
  \end{center}
}]
\makeatother

\section{LRP evaluation result for AI-TOD-V2}
\label{appendix: A}
To comprehensively evaluate the performance of Dome-DETR on AI-TOD-V2 \cite{xu2022detecting}, we adopt the Localization Recall Precision (LRP) metric \cite{2009.13592, faster-coco-eval-aitod}, which provides a fine-grained assessment of detection quality by jointly considering localization, recall, and precision. Unlike traditional metrics such as mAP, LRP explicitly accounts for errors in object localization, classification, and duplicate detections.

The LRP score is computed as follows:
\begin{equation}
    \text{LRP} = \frac{1}{N} \sum_{i=1}^{N} \left( \lambda_{\text{loc}} \cdot E_{\text{loc}, i} + \lambda_{\text{rec}} \cdot E_{\text{rec}, i} + \lambda_{\text{prec}} \cdot E_{\text{prec}, i} \right),
\end{equation}
where \( E_{\text{loc}, i} \) represents the localization error, which measures the IoU deviation between predicted and ground-truth bounding boxes. \( E_{\text{rec}, i} \) denotes the recall error, penalizing missed detections. \( E_{\text{prec}, i} \) quantifies the precision error, accounting for false positives and duplicate predictions. \( \lambda_{\text{loc}}, \lambda_{\text{rec}}, \lambda_{\text{prec}} \) are weighting factors to balance these error components.

For AI-TOD-V2, we report the Optimal LRP (oLRP), which is the minimum LRP score achieved when setting the optimal confidence threshold. Lower oLRP values indicate better detection performance. Results can be seen in Table \ref{tab:lrp_results}.

\section{Efficiency analysis of MWAS and PAQI}
\label{appendix: B}
This section presents an analysis of the efficiency of MWAS and PAQI across various detection scenarios, employing both quantitative and qualitative methods.

\subsection{Theoretical Limit of GFLOPs}
Thanks to its dynamic design, the computational cost of Dome-DETR adapts to the detection scenario. In dense scenes, the overhead increases, while in sparse scenes, it remains close to the baseline. This is attributed to the dynamic nature of \textbf{MWAS} and \textbf{PAQI}, which adjust feature enhancement and query initialization based on scene complexity.

In the densest scenarios, MWAS enhances all regions and PAQI generates the maximum number of queries. In contrast, under sparse conditions, only one region is enhanced and the minimum number of queries is used. These behaviors are based on experiments conducted on the AI-TOD-V2 dataset.

As shown in Table~\ref{tab:dome_detr_complexity}, Dome-DETR’s complexity varies significantly across scenarios. Under dense settings, the number of enhanced windows (100) and queries (1500) leads to higher GFLOPs: 193.8 (S), 282.8 (M), and 398.9 (L). In sparse scenes, both values drop (1 window, 300 queries), reducing GFLOPs by 36.4\%, 21.4\%, and 15.2\% for Dome-DETR-S, M, and L, respectively. This confirms the architecture’s scalability and efficiency across varying densities.

\input{tables/lrp_main}

\input{tables/computational_complexity_analysis}

\input{tables/paqi_analysis}

\subsection{Efficiency Analysis of MWAS}

\begin{figure}[t]
  \centering
  \includegraphics[width=\linewidth]{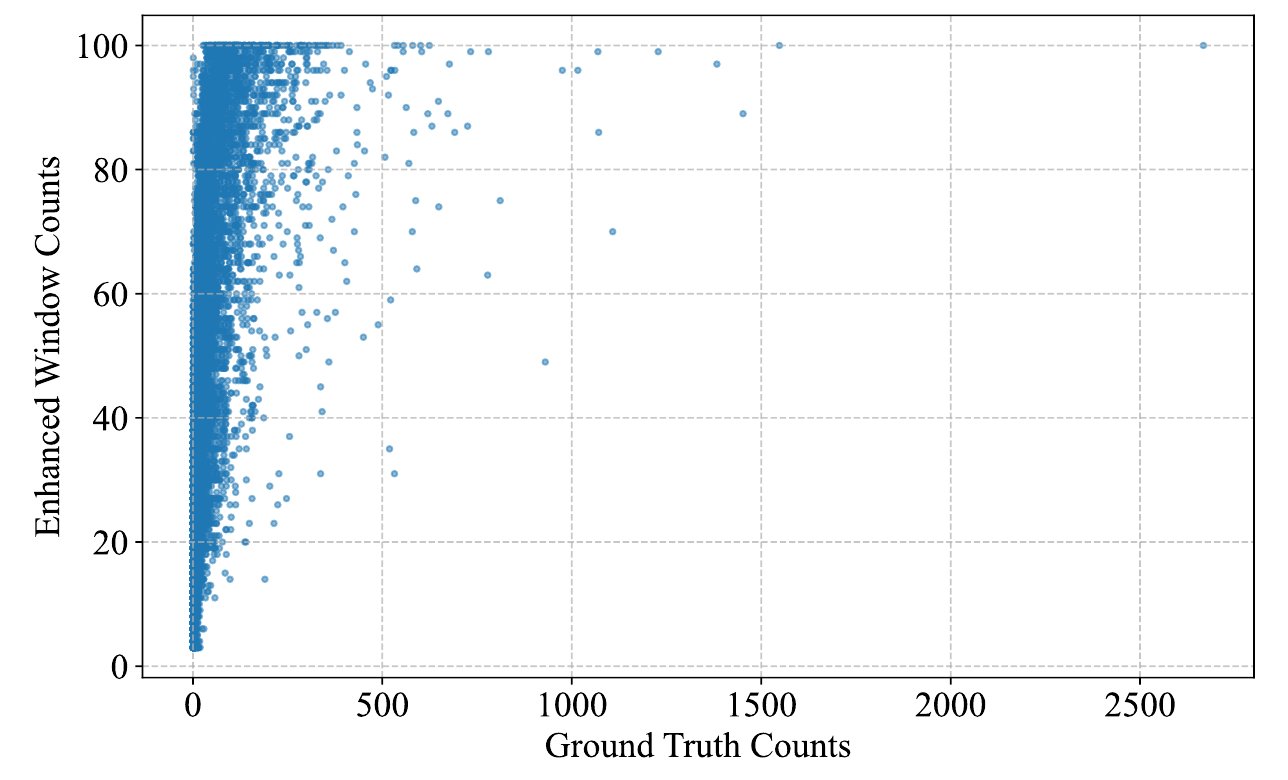}
  \caption{Ground truth counts and corresponding enhanced window counts of images across AI-TOD-V2 \textit{test} set of Dome-DETR-L.}
  \label{fig:window counts}
\end{figure}

To evaluate the efficiency of MWAS, we count the number of ground truth objects per image, \(N_{GT}\), and the number of enhanced regions by the MWAS module, \(N_{enhance}\), when running the Dome-DETR-L model on the AI-TOD-V2 benchmark's validation split. The corresponding plot is shown in Figure \ref{fig:window counts}. As observed from the figure, \(N_{enhance}\) exhibits an increasing trend as \(N_{GT}\) increases. Across the entire dataset, the average value is 36.1, with a standard deviation of 26.3, indicating that MWAS effectively sparsifies the attention on shallow features, thereby improving the model's efficiency.

\subsection{Efficiency Analysis of PAQI}

\begin{figure}[t]
  \centering
  \includegraphics[width=\linewidth]{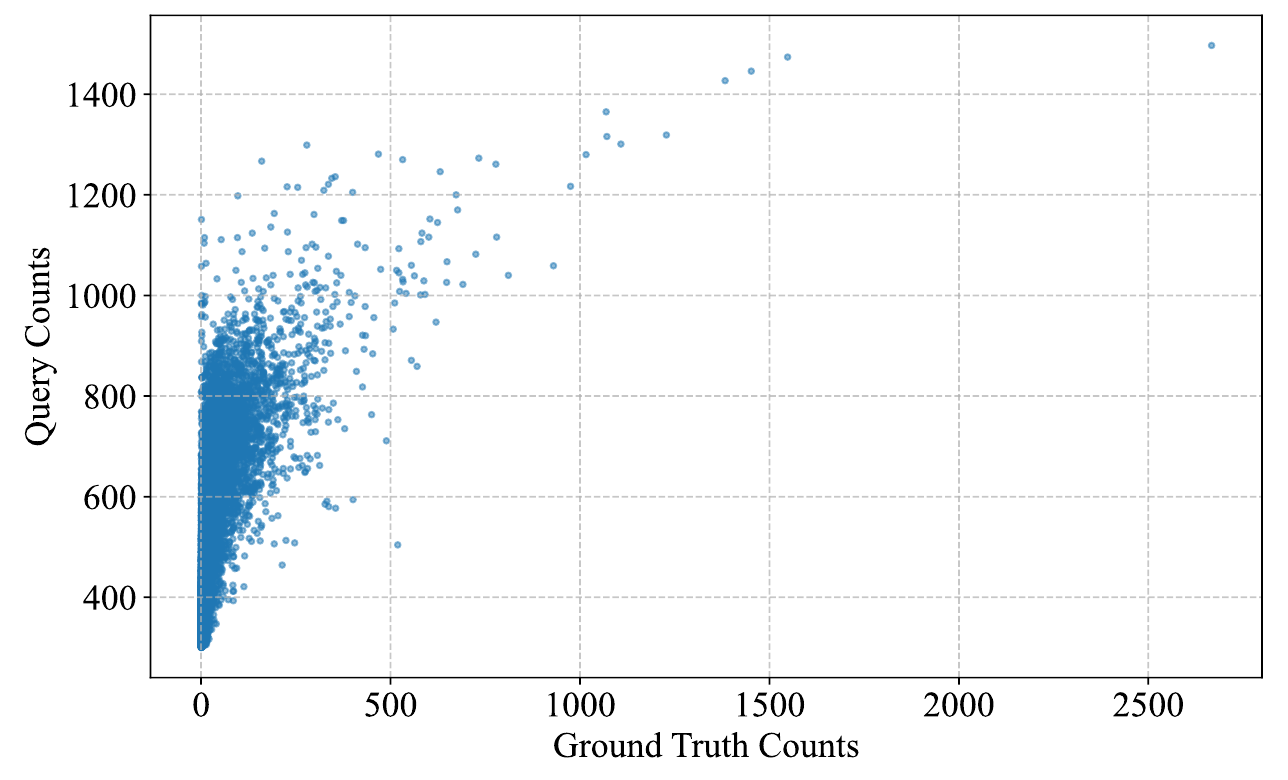}
  \caption{Ground truth counts and corresponding query counts of images across AI-TOD-V2 \textit{test} set of Dome-DETR-L.}
  \label{fig:query counts}
\end{figure}

To evaluate the effectiveness of PAQI, we count the number of ground truth objects per image, \(N_{GT}\), and the final number of initialized queries by the PAQI module, \(N_{Q}\), when running the Dome-DETR-L model on the AI-TOD-V2 benchmark's validation split. The corresponding plot is shown in Figure \ref{fig:query counts}. As observed from the figure, \(N_{Q}\) exhibits an increasing trend as \(N_{GT}\) increases, indicating that PAQI adapts its query allocation based on scene complexity. Across the entire dataset, the average value \(\hat{N_{Q}}\) is 498.9, with a standard deviation of 170.5, reflecting considerable flexibility. 

Furthermore, we introduce a new evaluation metric named \textbf{Query Ample Rate (QAR)}, which quantifies the probability that the number of initialized queries exceeds the number of ground truth objects in a dataset. Formally, we define QAR as:

\begin{equation}
\text{QAR} = \frac{1}{M} \sum_{i=1}^{M} \mathbbm{1} (N_{Q}^{(i)} > N_{GT}^{(i)}),
\end{equation}
where \( M \) is the total number of images in the dataset, \( N_{Q}^{(i)} \) and \( N_{GT}^{(i)} \) denote the number of initialized queries and ground truth objects in the \( i \)-th image, respectively, and \( \mathbbm{1}(\cdot) \) is the indicator function that returns 1 if the condition is true and 0 otherwise. A higher QAR indicates that the PAQI module frequently initializes more queries than ground truth objects, ensuring sufficient representation for accurate and robust object detection. A detailed comparison with other dynamic query quantity models is presented in Table~\ref{tab:paqi_comparison}. This indicates that PAQI effectively initializes queries based on the underlying object density distribution of each scene.

\section{Visualization of our proposed method}
\label{appendix: C}
\begin{figure*}[t]
    \centering
    \begin{subfigure}{0.8\linewidth}
        \centering
        \includegraphics[width=\linewidth]{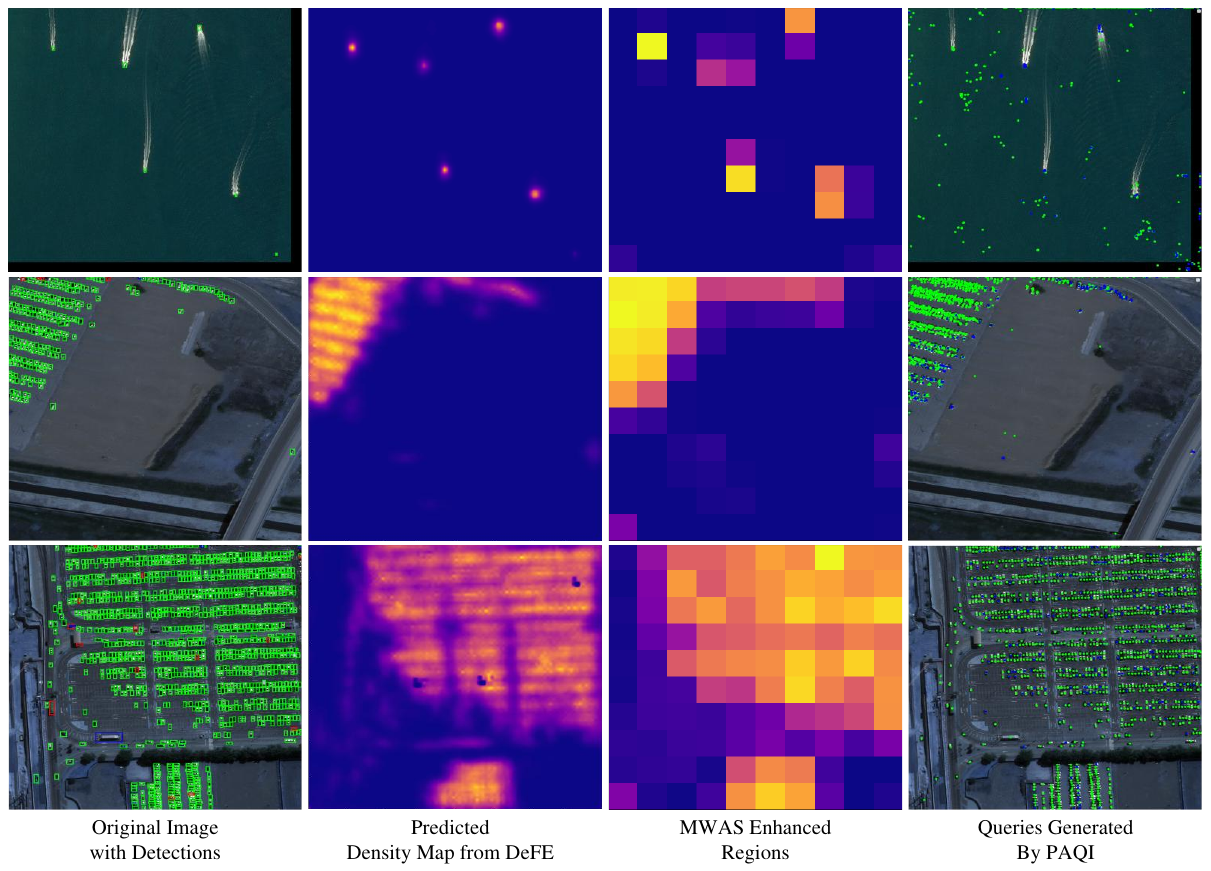}
        \caption{AI-TOD-V2 \textit{test} split}
        \label{fig:visual:aitod}
    \end{subfigure}
    
    \vspace{0.5em}
    
    \begin{subfigure}{0.8\linewidth}
        \centering
        \includegraphics[width=\linewidth]{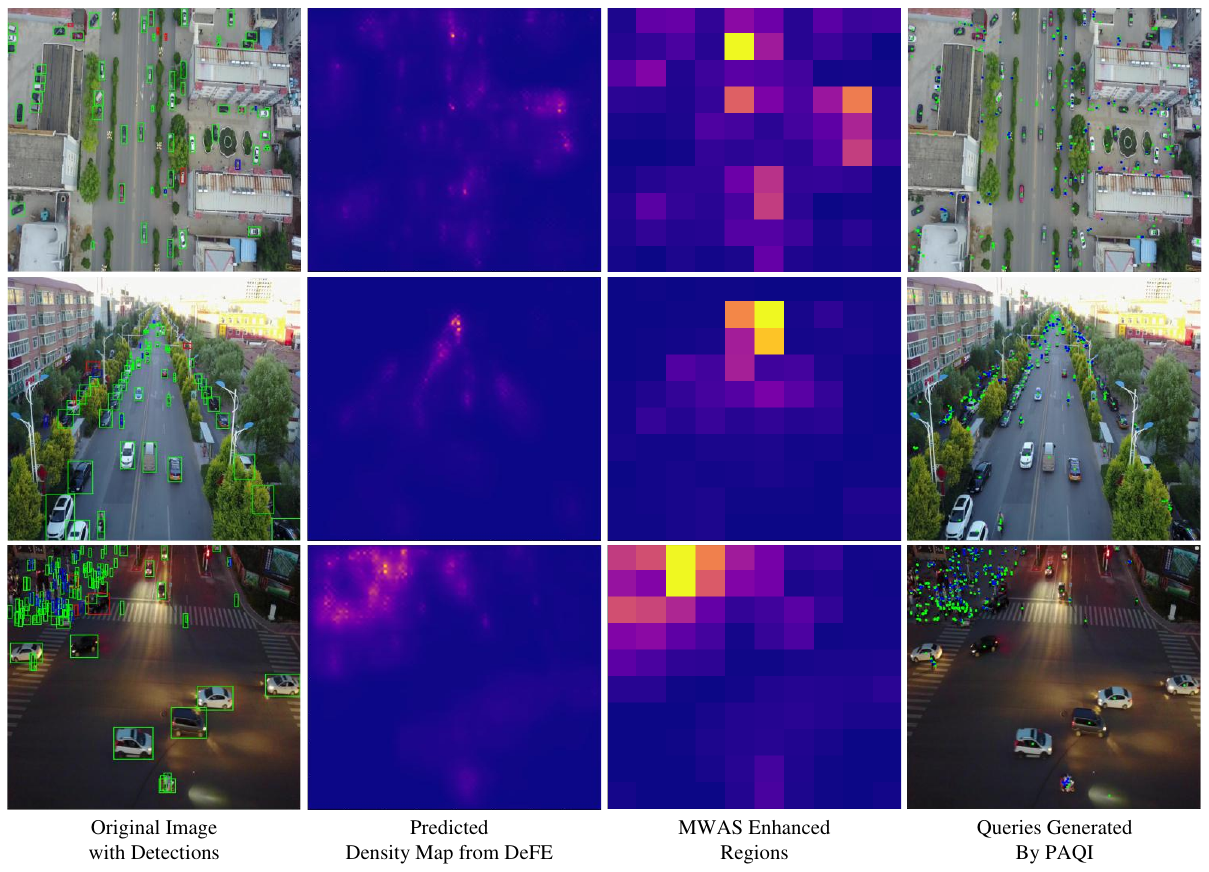}
        \caption{VisDrone \textit{val} split}
        \label{fig:visual:visdrone}
    \end{subfigure}
    
    \caption{Visualization of our proposed method under different scenes across two datasets.}
    \label{fig:visual:both}
\end{figure*}

Visualization of detection process of our proposed method can be seen in Figure \ref{fig:visual:aitod} and Figure \ref{fig:visual:visdrone}.

\textbf{Original Image with Detections.} The leftmost column displays original aerial images with detections. Green boxes denote true positives (TP) and red boxes indicate false positives (FP). Dome-DETR effectively detects tiny objects (e.g., vehicles, pedestrians) across varying object densities, demonstrating robustness in both sparse and dense scenes.

\textbf{Predicted Density Map from DeFE.} The second column shows density maps generated by the \textbf{Density-Focal Extractor (DeFE)}, encoding instance locations and relative scales. Brighter regions represent higher object density. These maps align well with ground-truth annotations, highlighting DeFE’s effectiveness in capturing spatial density.

\textbf{MWAS Enhanced Regions.} The third column illustrates regions enhanced by \textbf{Masked Window Attention Sparsification (MWAS)}. Guided by density maps, MWAS prunes background tokens and focuses on foreground regions, reducing computation while preserving object details. Axis-permuted attention further improves cross-window information flow.

\textbf{Queries Generated by PAQI.} The rightmost column presents queries from \textbf{Progressive Adaptive Query Initialization (PAQI)}. Green points are final queries; blue ones are filtered. PAQI adjusts query density based on scene complexity, enhancing recall in dense areas while maintaining efficiency in sparse regions.

The visualization confirms that the combination of DeFE, MWAS, and PAQI enables the model to achieve state-of-the-art performance while maintaining low computational costs.

\section{Implementation Details}
\label{appendix: D}

Table~\ref{tab:domdetr-settings} summarizes the hyperparameter settings for Dome-DETR. All variants use HGNetV2 backbones (B4, B2, B0 for L, M, S, respectively) pre-trained on ImageNet, and adopt the AdamW optimizer, following the baseline D-FINE \cite{peng2024d} setup.

Each model is trained with a total batch size of 8 for 120 epochs with advanced augmentation, followed by 40 epochs without it. All models use four feature levels for multi-scale representation, with feature extraction from encoder layer 3.

Dome-DETR-L and Dome-DETR-M use a GELAN hidden dimension of 128, while Dome-DETR-S uses 64. The transformer decoder consists of 6, 4, and 3 layers for Dome-DETR-L, M, and S, respectively. Training employs classification, regression, and $\mathcal{L}_{DRFL}$ loss (weight=1). The base learning rate is 2.5e-4 for Dome-DETR-L and 2e-4 for the others, with weight decay set at 1.25e-4 for Dome-DETR-L and 1e-4 for the rest. For stable training, $T_{init}$ is 5e-2 across all models. IoU thresholds are set at 0.4 ($IoU_{N}$) and 0.9 ($IoU_{M}$). The MWAS module uses a fixed window size of 10, and the adaptive position encoding (APE) module applies a single attention layer twice ($1 \times 2$) for efficient spatial encoding.

\begin{table*}[t]
    \centering
    \caption{Additional comparisons with classic works on VisDrone validation split. Latency is measured on an NVIDIA T4 GPU unless otherwise noted. Results show our proposed Dome-DETR achieve comparable performance with much lower latency.}
    \begin{tabular}{lccccccc}
    \toprule
    Method & Input Size $\downarrow$ & \#Params. $\downarrow$ & GFLOPs $\downarrow$ & Latency (ms) $\downarrow$ & AP (\%) $\uparrow$ & AP50 (\%) $\uparrow$ & AP75 (\%) $\uparrow$ \\
    \midrule
    Focus\&Detect \cite{koyun2022focus} & 1000$\times$600 & - & - & 1362 (NVIDIA 2080Ti) & \textbf{42.0} & 66.1 & \textbf{44.6} \\
    YOLC \cite{liu2024yolc} & 1024$\times$640 & - & \textbf{151} & 441 (NVIDIA 2080Ti) & 39.6 & 63.7 & 41.6 \\
    UFPMP-Det-ResNet50 \cite{huang2022ufpmp} & 1333$\times$800 & 65M & 443 & >100 & 36.6 & 62.4 & 36.7 \\
    UFPMP-Det-ResNeXt101 \cite{huang2022ufpmp} & 1333$\times$800 & 104M & 758 & >200 & 40.1 & \textbf{66.8} & 41.3 \\
    Dome-DETR-M (Ours) & 800$\times$800 & \textbf{24M} & 285 & \textbf{<15} & 36.1 & 59.8 & 39.4 \\
    Dome-DETR-L (Ours) & 800$\times$800 & 36M & 376 & \textbf{<20} & 39.0 & 61.1 & 40.8 \\
    Dome-DETR-L (Ours) & 1200$\times$800 & 36M & 402 & \textbf{<20} & 40.5 & 64.5 & 42.5 \\
    \bottomrule
    \end{tabular}
    \label{tab:additional_comparisons_visdrone}
\end{table*}

\section{Frequently Asked Questions During Review}
\label{appendix: E}

\subsection{Design Details of DeFE}

The Density-Focal Extractor is a lightweight yet effective module designed to generate precise object density priors, which guide subsequent feature enhancement and query initialization. Its key design details and innovations are as follows:

\textbf{Multi-scale dilated convolutions with strategic rates:} DeFE employs three parallel dilated convolutions with rates of 1, 2, and 3. This design is purposeful: a dilation rate of 1 captures fine-grained local features (critical for tiny object boundaries), while rates of 2 and 3 expand the receptive field to aggregate contextual information from medium and larger ranges, respectively. By fusing these multi-scale features, DeFE effectively models both the detailed texture of tiny objects and their spatial distribution patterns in complex scenes (e.g., dense clusters or sparse arrangements).

\textbf{End-to-end density probability generation:} After feature fusion, a global average pooling layer aggregates high-level contextual cues, ensuring the density estimation is not biased by local noise. A 1×1 convolution followed by a sigmoid activation then compresses the fused features into a pixel-wise density probability map (range 0–1), where higher values indicate regions with a higher likelihood of containing objects. This map serves as a "spatial attention guide" for subsequent modules, enabling the model to focus on foreground regions.

\textbf{Efficiency optimization:} Despite its multi-scale design, DeFE remains lightweight, adding only 0.8M parameters and 17.6 GFLOPs to the overall model. This efficiency is achieved through the use of depthwise separable convolutions (DS-conv) in the feature extraction stage, which drastically reduce computational overhead compared to standard convolutions, while bilinear upsampling preserves spatial resolution without excessive cost. This balance of performance and efficiency makes DeFE suitable for real-time applications.

\subsection{Difference Between Previous Cluster-aware Cropping Based Methods}

Some reviewers suggested comparing our method with classic approaches in the field of UAV object detection, such as Focus\&Detect, YOLC, and UFPMP. These methods mainly rely on the Cluster-aware Cropping strategy: they first cluster and crop regions of interest in the image, then perform detection using detectors, and finally reorganize and map the detection boxes back to the original image. Although these methods have unique advantages, they have an upper limit in real-time performance. This is because their strategies introduce heavy CPU and I/O tasks, making it impossible to achieve good parallelism.

In contrast, Dome-DETR emphasizes using density maps to guide the enhancement of regional features and the progressive initialization of Object Queries, and completes object detection in the image in \textbf{a single step}, thereby ensuring high detection efficiency and low latency.

More comparisons with these methods are shown in Table \ref{tab:additional_comparisons_visdrone}.

\subsection{Contribution of Each Component in PAQI}

To clarify the specific contribution of each sub-module in the Progressive Adaptive Query Initialization (PAQI) framework, we conducted ablation experiments by incrementally adding components to the baseline model:

\textbf{Candidate Query Selection:} This component expands the initial query budget from 300 to 1500, significantly increasing the model’s capacity to capture potential object instances. This expansion directly boosts recall (AR improves from 46.0 to 49.1) by covering more spatial regions, especially in dense scenes where tiny objects are clustered. However, the larger query pool also introduces higher computational overhead, as more queries require additional feature interactions.

\textbf{Query Splitting \& Density-Aware Sampling:} Building on the expanded query set, this module refines query allocation by distinguishing between "core queries" (targeting high-density object regions) and "flexible queries" (adapting to sparse areas). It suppresses redundant queries in background or low-information regions, reducing unnecessary computations (lower GFLOPs compared to the previous step). This refinement yields a notable AP improvement from 32.4 to 33.0, striking a better balance between coverage and efficiency.

\textbf{Bounding Box Generation \& Dynamic NMS:} The final component introduces density-aware non-maximum suppression (NMS), which adjusts suppression thresholds based on local object density. In dense regions with overlapping boxes, it retains more true positives by relaxing the threshold, while in sparse regions, it tightens the threshold to filter false positives. This dynamic mechanism pushes AP to the highest value of 33.3, ensuring optimal accuracy-efficiency trade-offs, particularly in complex scenes with varying object distributions. distributions.

\clearpage

\clearpage

\begin{table*}[t]
    \vspace{6cm}
    \centering
    \tabcolsep=0.5cm
    \caption{Hyperparameter settings for different Dome-DETR models.}
    \begin{tabular}{lccc}
        \toprule
        \textbf{Setting} & \textbf{Dome-DETR-L} & \textbf{Dome-DETR-M} & \textbf{Dome-DETR-S} \\
        \midrule
        Backbone Name & HGNetv2-B4 & HGNetv2-B2 & HGNetv2-B0 \\
        GELAN Hidden Dimension & 128 & 128 & 64 \\
        GELAN Depth & 3 & 2 & 1 \\
        Decoder Layers & 6 & 4 & 3 \\
        Sampling Point Number & (S: 4, M: 4, L: 4, X: 4) & (S: 4, M: 4, L: 4, X: 4) & (S: 4, M: 4, L: 4, X: 4) \\
        Number of feature levels & 4 & 4 & 4 \\
        Use encoder layer index & 3 & 3 & 3 \\
        Weight of $\mathcal{L}_{DRFL}$ & 1 & 1 & 1 \\
        Window Size of MWAS & 10 & 10 & 10 \\
        Number of attention layers in APE & $1 (\times 2)$ & $1 (\times 2)$ & $1 (\times 2)$ \\
        $T_{init}$ & 5e-2 & 5e-2 & 5e-2 \\
        $IoU_{N}$, $IoU_{M}$ & 0.4, 0.9 & 0.4, 0.9 & 0.4, 0.9 \\
        Base LR & 2.5e-4 & 2e-4 & 2e-4 \\
        Backbone LR & 1.25e-5 & 2e-5 & 1e-4 \\
        Weight Decay & 1.25e-4 & 1e-4 & 1e-4 \\
        Total Batch Size & 8 & 8 & 8 \\
        Epochs (w/ + w/o Adv. Aug.) & 120 + 40 & 120 + 40 & 120 + 40 \\
        \bottomrule
    \end{tabular}
    \label{tab:domdetr-settings}
\end{table*}




%% file: tables/lrp_main.tex
\begin{table}[!t]
    \centering
    \tabcolsep=0.23cm
    \caption{Comparison of LRP metrics on AI-TOD-V2 dataset. Lower values indicate better performance.}
    \begin{tabular}{lccc}
        \toprule
        \textbf{Method} & \textbf{oLRP} $\downarrow$ & \textbf{LRP FP} $\downarrow$ & \textbf{LRP FN} $\downarrow$ \\
        \midrule
        DINO-DETR \cite{zhang2022dino}& 78.9 & 27.5 & 35.3 \\
        DQ-DETR \cite{huang2024dq}& 73.4 & 25.1 & 33.5 \\
        D-FINE-L \cite{peng2024d} & 72.5 & 24.2 & 32.7 \\
        \rowcolor[HTML]{F5F5F5} \textbf{Dome-DETR-L (Ours)} & \textbf{70.1} & \textbf{23.1} & \textbf{32.2} \\
        \bottomrule
    \end{tabular}
    \label{tab:lrp_results}
\end{table}

%% file: tables/computational_complexity_analysis.tex
\begin{table}[t]
    \centering
    \caption{Computational Complexity of Dome-DETR Models in Different Scenarios}
    \label{tab:dome_detr_complexity}
    \begin{tabular}{lcccc}
        \toprule
        \textbf{Model} & \textbf{Scenario} & \textbf{\makecell[c]{Enhanced \\ Windows}} & \textbf{\makecell[c]{Query \\ Count}} & \textbf{GFLOPs} \\
        \midrule
        \multirow{2}{*}{Dome-DETR-S} & Dense & 100 & 1500 & 193.8 \\
                                      & Sparse & 1 & 300 & 123.3 \\
        \multirow{2}{*}{Dome-DETR-M} & Dense & 100 & 1500 & 282.8 \\
                                      & Sparse & 1 & 300 & 222.3 \\
        \multirow{2}{*}{Dome-DETR-L} & Dense & 100 & 1500 & 398.9 \\
                                      & Sparse & 1 & 300 & 338.4 \\
        \bottomrule
    \end{tabular}
\end{table}

%% file: tables/paqi_analysis.tex
\begin{table}[t]
    \centering
    \renewcommand{\arraystretch}{1.2}
    \caption{Comparison of dynamic query quantity methods on the AI-TOD-V2 \textit{test} set. \(\hat{N_{Q}}\) and \(\sigma(N_{Q})\) represent the mean and standard deviation of initialized queries, respectively. QAR is the Query Ample Rate.}
    \begin{tabular}{lccc}
        \toprule
        \textbf{Model} & \(\hat{N_{Q}}\) & \(\sigma(N_{Q})\) & \textbf{QAR} $\uparrow$ \\
        \midrule
        DQ-DETR \cite{huang2024dq} & 409.5 & 117.7 & 92.4 \\
        \rowcolor[HTML]{F5F5F5} \textbf{Dome-DETR-L (Ours)} & \textbf{498.9} & \textbf{170.5} & \textbf{99.9} \\
        \bottomrule
    \end{tabular}
    \label{tab:paqi_comparison}
\end{table}